\documentclass[final]{cvpr}

\usepackage{times}
\usepackage{epsfig}
\usepackage{graphicx}
\usepackage{amsmath}
\usepackage{amssymb}

\usepackage{xfrac}
\usepackage{subcaption}
\usepackage[export]{adjustbox}
\usepackage[table]{xcolor}
\usepackage{color}
\usepackage{rotating}
\usepackage{array}
\usepackage{fixltx2e}
\usepackage{xr}
\usepackage{wrapfig}
\makeatletter
\@namedef{ver@everyshi.sty}{}
\makeatother
\usepackage{tikz}

\definecolor{myblue}{rgb}{.1,0.3,0.9}
\definecolor{rowblue}{RGB}{220,230,240}%{230,240,250}
\usepackage{booktabs}
\usepackage{tablefootnote}
\usepackage{colortbl}
\usepackage{tabularx}
\usepackage{multirow}
\usepackage[margin=0cm,font=small,labelfont=bf]{caption}
\usepackage{xr}
\usepackage[scientific-notation=true]{siunitx}

\usepackage[pagebackref=true,breaklinks=true,colorlinks,bookmarks=false]{hyperref}

\newcommand{\new}[1]{{\textcolor{black}{#1}}}
\def\cvprPaperID{4368} % *** Enter the CVPR Paper ID here
\def\confYear{CVPR 2022}

\begin{document}
	
	%%%%%%%%% TITLE
	\title{Neural Point Light Fields}
	
	\author{Julian Ost$^{1}$\qquad Issam Laradji$^{2}$\qquad Alejandro Newell$^{3}$\qquad Yuval Bahat$^{3}$\qquad Felix Heide $^{1, 3}$\vspace{5pt}\\
		$^1$Algolux   \quad $^2$McGill \quad $^3$Princeton University}
	
	\maketitle
	
	%\footnotetext[1]{Shared contribution. See Author Contributions.}

% formatting stuff

\definecolor{Gray}{rgb}{0.5,0.5,0.5}
\definecolor{darkblue}{rgb}{0,0,0.7}
\definecolor{orange}{rgb}{1,.5,0} % something readable but different from todo
\definecolor{red}{rgb}{1,0,0} % something readable but different from todo

% taken from https://designnavigator.daimler.com/Daimler_Color_System
\definecolor{dai_ligth_grey}{RGB}{230,230,230}
\definecolor{dai_ligth_grey20K}{RGB}{200,200,200}
\definecolor{dai_ligth_grey40K}{RGB}{158,158,158}
\definecolor{dai_ligth_grey60K}{RGB}{112,112,112}
\definecolor{dai_ligth_grey80K}{RGB}{68,68,68}
\definecolor{dai_petrol}{RGB}{0,103,127}
\definecolor{dai_petrol20K}{RGB}{0,86,106}
\definecolor{dai_petrol40K}{RGB}{0,67,85}
\definecolor{dai_petrol80}{RGB}{0,122,147}
\definecolor{dai_petrol60}{RGB}{80,151,171}
\definecolor{dai_petrol40}{RGB}{121,174,191}
\definecolor{dai_petrol20}{RGB}{166,202,216}
\definecolor{dai_deepred}{RGB}{113,24,12}
\definecolor{dai_deepred20K}{RGB}{90,19,10}
\definecolor{dai_deepred40K}{RGB}{68,14,7}
%\definecolor{violettblau}{cmyk}{0.9, 0.6, 0, 0}
\definecolor{rot}{RGB}{238, 28 35}
\definecolor{apfelgruen}{RGB}{140, 198, 62}
%\definecolor{gelb}{RGB}{255, 229, 0}
\definecolor{orange}{RGB}{244, 111, 33}
\definecolor{pink}{RGB}{237, 0, 140}
\definecolor{lila}{RGB}{128, 10, 145}
%\definecolor{hellgrau}{RGB}{224, 224, 224}
%\definecolor{mittelgrau}{RGB}{128, 128, 128}
%\definecolor{dunkelgrau}{RGB}{80,80,80}
\definecolor{anthrazit}{RGB}{19, 31, 31}

\newcommand{\heading}[1]{\noindent\textbf{#1}}
\newcommand{\note}[1]{{\em{\textcolor{orange}{#1}}}}
\newcommand{\todo}[1]{{\textcolor{red}{\bf{TODO: #1}}}}
\newcommand{\comments}[1]{{\em{\textcolor{orange}{#1}}}}
\newcommand{\changed}[1]{#1}
\newcommand{\place}[1]{ \begin{itemize}\item\textcolor{darkblue}{#1}\end{itemize}}
\newcommand{\de}{\mathrm{d}}

\newcommand{\normlzd}[1]{{#1}^{\textrm{aligned}}}

% dimensions
\newcommand{\ttime}{\tau}               % time coordinate
\newcommand{\x}{\Vect{x}}               % spatial coordinates in vectorized form
\newcommand{\z}{z}               % depth coordinate of volume

\newcommand{\npixels}{n}               % num pixels
\newcommand{\ntime}{t}               % num timesteps

%image formation
\newcommand{\illfunc}     {g}
\newcommand{\pathfunc}     {s}
\newcommand{\camfunc}     {f}

% notation for image formation
\newcommand{\irradiance}{E}
\newcommand{\exposure}{b}
\newcommand{\pmdfunc}{f}                % modulation on the PMD side
\newcommand{\lightfunc}{g}              % modulation of the light
\newcommand{\period}{T}                 % temporal period of the modulation
\newcommand{\freqm}{\omega}                % frequency of modulation
\newcommand{\illphase}{\rho}             % frequency of modulation harmonics
\newcommand{\sensphase}{\psi}             % frequency of modulation harmonics of pmd camera
\newcommand{\pmdphase}{\phi}            % phase of PMD modulation
\newcommand{\omphi}{{\omega,\phi}}      % shorthand for freq/phase pair
\newcommand{\numperiod}{N}              % #periods integrated over
\newcommand{\att}{\alpha}               % geometric & photometric attenuation
\newcommand{\pathspace}{{\mathcal{P}}}  % space of all light paths

% regular modulation based cameras
\newcommand{\atan}{\operatorname{atan}}

% math stuff
\newcommand{\Fourier}{\mathfrak{{F}}}         % fourier transform
\newcommand{\conv}     {\otimes}
\newcommand{\corr}     {\star}
\newcommand{\Mat}[1]    {{\ensuremath{\mathbf{\uppercase{#1}}}}} %Matrix 
\newcommand{\Vect}[1]   {{\ensuremath{\mathbf{\lowercase{#1}}}}} %Vector
\newcommand{\Id}				{\mathbb{I}} %Identity matrix
\newcommand{\Diag}[1] 	{\operatorname{diag}\left({ #1 }\right)} %Diagonalized matrix
\newcommand{\Opt}[1] 	  {{#1}_{\text{opt}}} %Optimal point of an optimization
\newcommand{\CC}[1]			{{#1}^{*}} %Convex conjugate
\newcommand{\Op}[1]     {\Mat{#1}} %Operator
\newcommand{\minimize}[1] {\underset{{#1}}{\operatorname{argmin}} \: \: } %Minimize w.r.t.
\newcommand{\maximize}[1] {\underset{{#1}}{\operatorname{argmax}} \: \: } %Maximize w.r.t.  
\newcommand{\grad}      {\nabla}

% notation for method
\newcommand{\Basis}{\Mat{H}}         		% Matrix basis
\newcommand{\Corr}{\Mat{C}}             % measurement matrix
\newcommand{\correlem}{\bold{c}}             % measurement matrix element
\newcommand{\meas}{\Vect{b}}            % measurement vector
\newcommand{\Meas}{\Mat{B}}            % measurement matrix
\newcommand{\MeasNormalized}{\Mat{B}^{\textrm{new}}}            % measurement matrix
\newcommand{\Img}{H}                    % transient image
\newcommand{\img}{\Vect{h}}             % vectorized image
\newcommand{\latentresponse}{\alpha}

\newenvironment{customlegend}[1][]{%
        \begingroup
        % inits/clears the lists (which might be populated from previous
        % axes):
        \csname pgfplots@init@cleared@structures\endcsname
        \pgfplotsset{#1}%
    }{%
        % draws the legend:
        \csname pgfplots@createlegend\endcsname
        \endgroup
    }%

    % makes \addlegendimage available (typically only available within an
    % axis environment):
    \def\addlegendimage{\csname pgfplots@addlegendimage\endcsname}

	%\todo{teaser figure}
	%%%%%%%%% ABSTRACT
	\begin{abstract}
	We introduce Neural Point Light Fields that represent scenes implicitly with a light field living on a sparse point cloud. Combining differentiable volume rendering with learned implicit density representations has made it possible to synthesize photo-realistic images for novel views of small scenes. As neural volumetric rendering methods require dense sampling of the underlying functional scene representation, at hundreds of samples along a ray cast through the volume, they are fundamentally limited to small scenes with the same objects projected to hundreds of training views. Promoting sparse point clouds to neural implicit light fields allows us to represent large scenes effectively with only a single radiance evaluation per ray. These point light fields are as a function of the ray direction, and local point feature neighborhood, allowing us to interpolate the light field conditioned training images without dense object coverage and parallax. We assess the proposed method for novel view synthesis on large driving scenarios, where we synthesize realistic unseen views that existing implicit approaches fail to represent. We validate that Neural Point Light Fields make it possible to predict videos along unseen trajectories previously only feasible to generate by explicitly modeling the scene. 
	%We present Neural Point Light Fields that reconstructs a light field of a given scene from images and point clouds.
	%This novel approach exploits the point cloud data that has been captured together with images of the scene to embed features on those points and with that enable neural renderings of novel views. This allows us to apply the latest advances in neural rendering to large scenes evaluating the representation function one time per rendered ray. 
	%Highly desirable for realistic simulations.
	%
	%Our method first extracts and learns point features, that are directly embedded on the point cloud. For each ray of a virtual camera we extract a unique feature from the point cloud. This conditions a light field representation of the scene that renders an image.
	%
	%We compare the reconstruction and novel views against pure implicit and explicit representations as well as novel hybrid approaches on large automotive scenes. In various ablations, we show the effects of our design choices and the manipulation of ray direction and point clouds.
	%
	%With our method we present... 
	\end{abstract}
	
	%%%%%%%%%% BODY TEXT
	\vspace{-18pt}
	\section{Introduction}\label{sec:intro}
Learning implicit volumetric scene representations has made it possible to synthesize photo-realistic images of single scenes~\cite{lombardi2019neuralvolumes, mildenhall2020nerf, niemeyer2020dvr, sitzmann2019srns}. The most successful methods combine a conventional volumetric rendering approach with a coordinate-based neural network that predicts density and radiance~\cite{mildenhall2020nerf}. As such, instead of explicitly storing density and radiance in a high-dimensional 5D volume, these methods represent this volume as a learned function, that can be further decomposed into radiance and illumination~\cite{zhang2021nerfactor,srinivasan2021nerv,boss2021nerd}. Although the implicit volumetric representation is highly memory-efficient and differentiable, it also fundamentally requires sampling the volume, that is evaluating the coordinate-based network, hundreds of times for each ray for a given pixel. This mandates long training and small volumetric support inside the volume. %training volume. 

To tackle these challenges, hybrid representations \cite{hao2021gancraft, liu2020neural, hedman2021baking} are used to embed or ``bake'' local radiance functions on explicit sparse proxy representations such as coarse voxel grids, point clouds or meshes to enable faster rendering by ignoring empty space. While this approach drastically improves rendering speed at test time, it still requires volumetric sampling during training. This is because the scene geometry must be learned during the training process. These methods share the limitations of volumetric approaches during training and, as such, have also been limited to small scenes that are costly to train. Learning representations for large outdoor scenes is an open challenge.

\begin{figure}[t!]
    \includegraphics[width=\columnwidth, trim={0cm 0cm 0cm 0cm},clip]{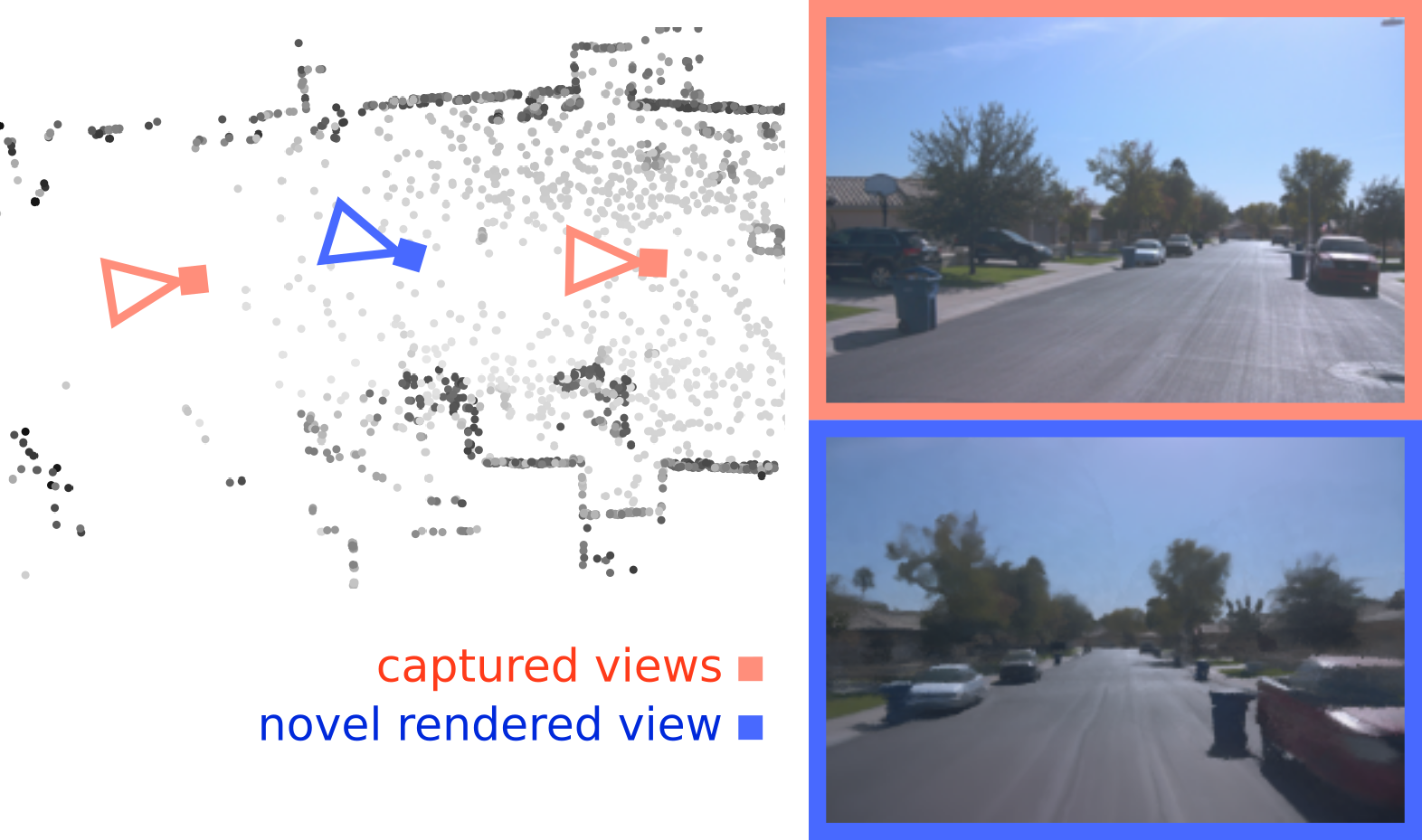}

	\caption{Neural Point Light Fields 
	encode
    % 	encodes
    the information of a Light Field representation of a scene on a point cloud capture. An image is rendered for each camera ray based on the local encoding of the Light Field on relevant points.}
	\vspace{-16pt}
\end{figure}

Unfortunately, approaches that are free of implicit representations do not yet offer an alternative. Specifically, explicitly storing features on proxy geometry \cite{riegler2021stable,riegler2020free, kopanas2021point} has not been able to achieve the same quality as volumetric methods when interpolating a view without a nearby training sample. Existing formulations utilize geometry as a projection canvas combined with features extracted from target views, and therefore require a large number of input images near the target view. 

In this work, we depart from volumetric models and introduce Neural Point Light Fields, a local implicit representation that encodes a light field on a point cloud. The proposed representation supports novel view synthesis in large outdoor scenes without strong parallax needed as in volumetric methods. Although recent automotive depth estimation networks make it possible to estimate dense depth point clouds from video data, we assume measured lidar point clouds as input to our method, especially as lidar data is readily available in most outdoor vehicle datasets~\cite{sun2020scalability, geiger2012kittivisonbenchmark} and recently released smartphones.
% cellphones.
Although sparse, the lidar geometry provides enough cues to encode a local light field on the point cloud. Instead of a 5D volumetric radiance function, or a conventional 4D light field~\cite{levoy1996light}, we propose to formulate a light field only depending on the two dimensional ray direction and a one dimensional index pointing to a point cloud feature%, an implicit 3D representation. 
This formulation makes it possible to evaluate a \emph{single} radiance prediction per ray.

We extract features for each point with a learned feature extractor on point cloud projections \cite{goyal2021revisiting}. For a given camera pose, we shoot rays for each pixel and select a set of close points inside the point cloud. The features from these selected points are then weighted by passing the points relative position to the ray and features through an attention module, resulting in a single ray feature code. The color for each ray is then reconstructed by an implicit light field representation conditioned by this feature code. %To enforce local feature variations per point we store the point features after that first training stage and refine the point features in a second training stage. 
We assess the proposed method on a large-scale automotive driving dataset~\cite{sun2020scalability} and demonstrate novel view synthesis along unseen trajectories with quality unseen before.

Specifically, we make the following contributions
\begin{itemize}
    \item We introduce Neural Point Light Fields, a representation that implicitly encodes features in a point cloud,
    requiring only a single radiance evaluation 
    % allowing for a single radiance evaluation 
    per ray.
    \item The proposed method lifts the restrictions of volumetric 
    % neural
    scene representations by exploiting sparse geometry available in estimated or captured point clouds.
    \item We validate the proposed method on novel video synthesis tasks for large-scale driving scenes, 
    demonstrating the proposed method's capability of generating realistic novel views along trajectories which cannot be handled by existing implicit representation methods.
    % where we demonstrate that the proposed method is capable of generating realistic novel views along trajectories that existing implicit scene representation methods fail for. 
\end{itemize}
Our code and trained models are available on our website: {\small \url{https://light.princeton.edu/neural-point-light-fields}}
%-------------------------------------------------------------------------
\paragraph{Scope}\label{sec:scope}
Even though existing automotive datasets include data from multiple cameras, lidar and radar sensors, we focus on learning from a single camera with a single trajectory per scene, and without highly dynamic scene motion. \new{In contrast to
densely observing the scene across a full hemisphere
% observations densely sampled on the hemisphere
~\cite{mildenhall2020nerf}, 
% in our case, captured images 
captured images in our case
are sparsely
% sampled
distributed
along the driving trajectory
%, often with only a handful of views per scene object.
} 
We note that extending training to
% training on 
multiple camera views is not straightforward, as camera poses, exposure and tone-mapping differences have to be accounted for.
% Incorporating
Exploiting 
multiple cameras and adding dynamic object support to the proposed method could
% be
constitute
exciting future directions.

%We solely trained on images of the front camera for two reasons. All methods requires a precise global camera pose for the entire scene and with respect to all cameras to not produce blurry predictions, which is not given from raw data. Images from different cameras in the data set show different exposure and white balance and thus produce inconsistent results for all methods. Both issues might be solved in future work using tone mapping and camera pose refinement modules.
	
	%------------------------------------------------------------------------
	\section{Related Work}\label{sec:related_work}
%-------------------------------------------------------------------------
\noindent\textbf{Novel View Synthesis.}
Synthesizing novel views from a set of unstructured images of a scene is a long standing problem in computer vision and graphics. Early work on image-based rendering introduced light fields \cite{levoy1996light} as a 4D parameterization of light rays and their respective radiance in a scene. Light fields are derived by considering a convex subspace of the 5D plenoptic function \cite{adelson1991plenoptic} that  parameterizes a ray by a point in space and a direction. Conventional light field rendering, i.e., interpolation of novel views, requires a large set of densely sampled views of the light field, as traditional optimization methods~\cite{wanner2012globally, wanner2013variational} handle only small parallax changes between the interpolated and measured view. Recently, methods relying on deep learning~\cite{mildenhall2019llff}
allowed recovering light field from plane sweep volumes, using 3D convolutional neural networks.
% introduced a method to recover the light field using from plane sweep volumes using 3D convolutional neural networks. 

An orthogonal line of work investigates the reconstruction of explicit 3D models from a set of images. By optimizing the reprojection error between features found in all images, multi-view reconstruction methods are capable of reconstructing the underlying scene geometry and camera poses \cite{argarwal2011romeinaday,schoenberger2016mvs}. These methods
% are able to 
can
reconstruct large scenes, but
% also
require many images to achieve high quality, and, in contrast to image-based rendering methods, 
% they
struggle to synthesize photorealistic novel views.

%-------------------------------------------------------------------------
\vspace{0.5em}\noindent\textbf{Neural Scene Representations.}
%
% A large body of work has emerged that
An emerging large body of work 
explores learned representations in scene reconstruction pipelines. These neural rendering approaches are able to generate photo-realistic novel views \cite{lombardi2019neuralvolumes, niemeyer2020dvr},
% and they are capable of 
while
reconstructing high-quality scene geometry. Existing methods rely on explicit, implicit, or hybrid representations of the scene. Explicit methods encode texture or radiance on recovered proxy 
scene geometry,
% geometry of the scene that hosts texture or radiance features, 
such as meshes \cite{thies2019deferred}, multi-planes \cite{flynn2019DeepView,lu2020layeredNeural,mildenhall2019llff,srinivasan2019viewextrapolationmultiplaneimages,zhou2018stereo}, voxels \cite{sitzmann2019deepvoxels} or points \cite{aliev2020neural,pittaluga2019revealing}. Instead of jointly recovering geometry and appearance,
% having to recover geometry and appearance jointly,
these methods can focus on
% the recovery of
recovering
image 
% detail
details. 
% However, at the same time,
Nonetheless,
relying on explicit proxy geometry 
% also
limits the achievable image quality.
To overcome the reliance on such geometry, researchers explored implicit representations 
% that represent scenes
using coordinate-based networks,
e.g., the successful NeRF method \cite{mildenhall2020nerf}.
% . The most successful methods \cite{mildenhall2020nerf} represent volumetric density using such a coordinate-network. 
However, 
% while
achieving photo-realistic quality for diverse tasks~\cite{martinbrualla2020nerfw, park2021nerfies, xian2021space, srinivasan2021nerv, Schwarz2020NEURIPS, Niemeyer2020GIRAFFE,ost2021neural}
% , this
comes at the cost of expensive training and testing. The lack of explicit geometric knowledge requires
densely evaluating the implicit network within the volume,
% evaluating the implicit network densely in the volume 
with the majority of samples located in empty space,
and therefore not contributing 
% that do not contribute 
to the rendered pixel color. Extensions \cite{garbin2021fastnerf} have tackled this issue at test time evaluation by either predicting the sampling regions \cite{neff2021donerf,arandjelovic2021nerf} or explicitly extracting proxy geometry \cite{liu2020neural} after training. DS-NeRF~\cite{deng2021depth} uses 3D keypoints reconstructed from COLMAP on a scene to supervise the opacity prediction with those sparse keypoints, which speeds up training. Neural Sparse Voxel Fields (NSVF)~\cite{liu2020neural} 
% are
use
a hybrid representation that stores implicit functions in a voxel grid. \new{NeRF++ proposes to separate background and foreground scene components~\cite{zhang2020nerf++}, which help improve the rendering quality, primarily for distant scene objects.} 
However, all
% All
of these methods \emph{struggle with large scale outdoor scenes or scenes with very few view 
directions}.
% observations}.
% \new{Separating background and foreground scene components~\cite{zhang2020nerf++} can only help improve the quality for distant scene objects.} 
In contrast, the proposed approach allows 
% for
rendering
large outdoor scenes from a sparse set of observations, by introducing a light field parameterization on sparse scene geometry. 

% \yb{I didn't comment that:}
%Gancraft doesn't apply
%With GANcraft \cite{hao2021gancraft} Hao et al. presented a method that generates photo realistic views from a previously constructed voxel world model encoding ray features on the voxels with an MLP. Ray features are fed into a CNN renderer. 
%Focusing on generative tasks from a known voxel representation the method does not suite the challenge of novel view synthesis from a captured scenes.

% \todo{Generalization} Methods that use features extracted from images and viewing direction either to condition a volumetric sampling point \cite{wang2021ibrnet, yu2021pixelnerf}. 
% \todo{add light field approaches to their category} \cite{liu2021nelf,sitzmann2021light}, Deep Surface Light Fields \cite{chen2018deepSLF}

%-------------------------------------------------------------------------
\vspace{0.5em}\noindent\textbf{Multi-View Structure (MVS) Reconstruction.}
Reconstructing geometry such as point clouds or meshes from images \cite{schoenberger2016mvs, schoenberger2016SfMrevisited} can guide the training of implicit scene representations~\cite{deng2021depth} or offer a scaffold for learned features~\cite{riegler2021stable, kopanas2021point}. Riegler and Koltun \cite{riegler2021stable, riegler2020free} propose such geometric scaffolds living on MVS-meshes. Kopanas et al. \cite{kopanas2021point} showed that optimizing point locations from an initial point cloud, together with their novel view synthesis pipeline, can compensate for 
% the
errors during reconstruction from MVS. These methods and similar \cite{aliev2020neural} point based approaches use point clouds as a geometric proxy, while following a strict rendering and projection approach. In contrast, we propose a method that uses features not only by projecting them on to a proxy geometry, but encodes them from a 3D point cloud, and requires no input images during test time.

\new{In the context of automotive scene reconstruction, SurfelGAN~\cite{yang2020surfelgan} proposes a representation with discrete textured 
surface elements
(surfels), recovered from captured Lidar and RGB data. Novel views are rendered by a generator network from projections of the surfel RGB data. In contrast, we learn features directly embedded in the captured point cloud.}

Encoding features directly on a point clouds has been extensively explored \cite{qi2017pointnet} for diverse tasks. Recent work revisited the use of multi-view projections of a point cloud for classification tasks \cite{goyal2021revisiting, hamdi2021mvtn}, similar to the proposed reconstructions from point clouds, but
without
% not
using image features. Their method is robust to occlusions \cite{hamdi2021mvtn}, and achieves state-of-the-art results on
% their
downstream tasks. 
Rather than
% Instead of
solving a classification or segmentation task, we show that multi-view point cloud encoding can deliver rich local point features for reconstruction of novel views.
	%------------------------------------------------------------------------
	\section{Point Light Fields}\label{sec:method}
In this section, we introduce Point Light Fields. A Point Light Field encodes the light field of a scene on sparse point clouds. Assuming a camera-lidar sensor setup typical in robotic and automotive contexts~\cite{geiger2012kittivisonbenchmark}, at time step $i$, the proposed method learns an RGB frame $\mathcal{I}_i$ as input and the corresponding point cloud capture $\mathcal{P}_i$. To learn a light field embedded on the point clouds corresponding to a video sequence, we devise three steps: an encoding step, a feature aggregation, and a point-conditioned light field prediction, all of which we describe in the following.
% Given a set of point cloud captures $\mathcal{P}$ the method learns to reconstruct a Light Field from a local encoding of that Light Field. 
\begin{figure*}[htb]
\vspace{-15pt}
	\centering
	\includegraphics[width=0.98\textwidth]{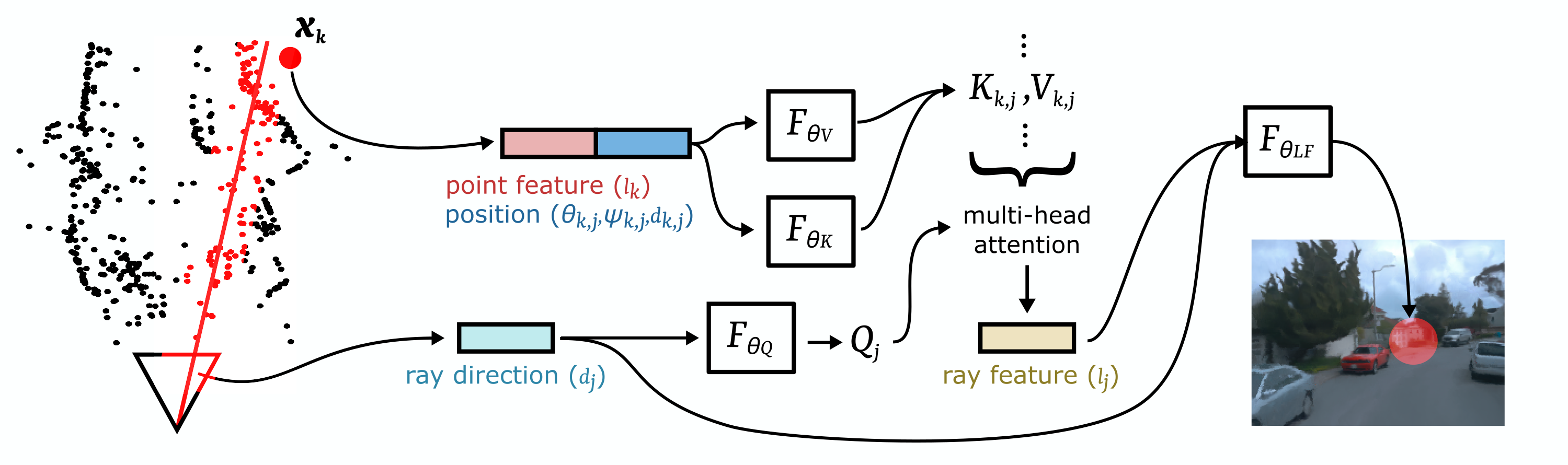}
	\vspace{-10pt}
	\caption{Neural Point Light Field Rendering Pipeline. For each ray $\boldsymbol{r}_j$, a set of $K$ closest points is selected from a point cloud of the scene. From each point $\boldsymbol{x}_k$, a feature vector $\boldsymbol{l}_k$ and the relative location with respect to $\boldsymbol{r}_j$ predict a key and value vectors. The most relevant point features are aggregated for the ray with a multi-head attention module, using the encoded ray direction $\boldsymbol{d}_j$ to form the query vector. A light field function $F_{\theta LF}$ computes the ray color given the ray feature $\boldsymbol{l}_j$ and ray direction $\boldsymbol{d}_j$.} \label{fig:overview}
	\vspace{-10pt}
\end{figure*}

%-------------------------------------------------------------------------
\subsection{Per-point Feature Encoding}
We first produce a feature embedding for each point in the point cloud. To do this, we follow the strategy presented by Goyal et al. \cite{goyal2021revisiting}. The input point cloud is projected onto six planes, producing sparse depth images. These images are each fed directly into a convolutional network. We use the initial layers of a vanilla ResNet18 \cite{he2016deep} to extract per-pixel features at one-quarter the input resolution. For a given point $\boldsymbol{x}_k$, we retrieve the corresponding feature vector at its projected location in each of the six views. These are concatenated together to produce the final feature encoding  $\boldsymbol{l}_k \in \mathbb{R}^{6 \times 128}$.

We find it sufficient to normalize input point clouds to a canonical cube bounded by $[-1, 1]$ and use the 6 sides of the cube as projection planes. This works robustly even given the complexity of in-the-wild large-scale scenes. We perform ablations comparing features encoded using this strategy against alternative point-based models such as PointNet~\cite{qi2017pointnet}, see Supplementary Material.

The learned per-point features $\boldsymbol{l}_k$ do not depend on any image data and can be trained end-to-end with the full light field rendering. We can introduce augmentations such that the model does not overfit to a particular arrangement of points. This includes sampling different subsets of points from the full captured point cloud, and using point cloud captures from nearby time steps.

\subsection{Light Field Feature Interpolation}\label{ssec:ray_attn}
Given a set of points $\mathcal{P}_{i} = \{\boldsymbol{x}_0, ..., \boldsymbol{x}_N\}_i \text{with } \boldsymbol{x}_k\in \mathbb{R}^3$, their encoded features $\boldsymbol{l}_k \in \mathbb{R}^{6 \times 128}$, and a camera view $C_i$, defined by its intrinsic $\boldsymbol{K}$, extrinsic $\boldsymbol{E}_i$ and sensor dimensions $W$ and $H$, we aggregate the 
features that are relevant for reconstructing 
% relevant features that reconstruct
the local light field around each ray. For all $W \times H$ pixels from $C_i$ we cast a set of rays $\mathcal{R}_i$ into the scene using a pinhole camera model. Each $\boldsymbol{r}_j \in \mathcal{R}$ is defined by its origin $\boldsymbol{o}_j$ and viewing direction $\boldsymbol{d}_j$.

%-------------------------------------------------------------------------
\vspace{0.5em}\noindent\textbf{Local Point Selection.}
The local point cloud encoding can explain the scene properties at their sparse locations. 
Explicitly representing high-frequency light field details from all views would necessitate a dense descriptor.
% To explicitly represent high-frequency light field detail from all views a dense descriptor would be necessary. 
Instead, we implicitly interpolate a representation descriptor for each ray. The work of DeVries et al. \cite{devries2021unconstrained} shows that the interpolation of local latent descriptors allows for implicit scene representations
for
% of
large indoor scenes. Unlike their regular grid structure, we want to leverage the information given through the geometric properties of the point cloud.
We assume that point features $\boldsymbol{l}_k$ hold enough information not only to represent the light field at their exact location, but also in their neighbourhood. 
For each ray $\boldsymbol{r}_j$,
we aggregate a descriptor
% for each ray $\boldsymbol{r}_j$
from a relevant set of sparse points. To this end, we select a set of $K$ points $P_{j,i} \subset \mathcal{P}_{i}$ inside the viewing frustum of the camera $C_i$, with the smallest orthogonal distance $d_{k,j}$ between the points and the ray
% , as formulated in Eq.~\ref{eq:pt_ray_d}. 
\begin{equation}
    \cos{\left( \varphi_{k,j} \right)} = \boldsymbol{d}_{j,i} \cdot \left( \frac{\boldsymbol{x}_{k,i} - \boldsymbol{o}_{j,i}}{||\boldsymbol{x}_{k,i} - \boldsymbol{o}_{j,i}||} \right),
    \label{eq:phi}
\end{equation}
%
% \begin{equation}
%     sin\left( \varphi_{j,k} \right) = sin\left( arccos\left( cos\left( \varphi_{j,k} \right)\right) \right) = \sqrt{1 - cos\^2left( \varphi_{j,k} \right)}
% \end{equation}
%
\begin{equation}
    \begin{aligned}
        d_{k,j}=  \sin{\left( \varphi_{k,j} \right)} \cdot \left( \boldsymbol{x}_{k,i} - \boldsymbol{o}_{j,i} \right) & \\
        \text{with } \sin{\left( \varphi_{k,j} \right)} = \sqrt{1 - \cos{^2\left( \varphi_{k,j} \right)}}. &
    \end{aligned}    
\label{eq:pt_ray_d}
\end{equation}

The ray origin $\boldsymbol{o}_{j,i}$, 
% the
normalized ray direction $\boldsymbol{d}_{j,i}$, and 
% the
point $\boldsymbol{x}_{k, i}$ are all given in a local reference frame centered in the captured $P_{i}$.
A light field descriptor is then generated for each ray, considering all encoded features on the points in $P_{j,i}$. 

%-------------------------------------------------------------------------
\vspace{0.5em}\noindent\textbf{Ray-centric Point Encoding.}
There are several immediate choices for the point embeddings of $P_{j,i}$, including average pooling, max pooling or a linear weighting by the distance $d_{j,k}$ of the selected $K$ point features. However, these interpolation methods are ambiguous, 
i.e., they 
% thus
can deliver the same descriptor for various rays and features on the same set of closest points $P_{i, j}$. In order to ensure a consistent and unique description for each ray from the 
set $P_{j,i}$, we must use the
% same set, $P_{j,i}$ must consider an
unambiguous relative position of all points with respect to that ray, with coherence across different time steps $i$ of the same scene. 

\begin{figure}[t!]
	\centering
	\includegraphics[width=0.39\textwidth, trim={1.5cm 0cm 1.5cm 0cm},clip]{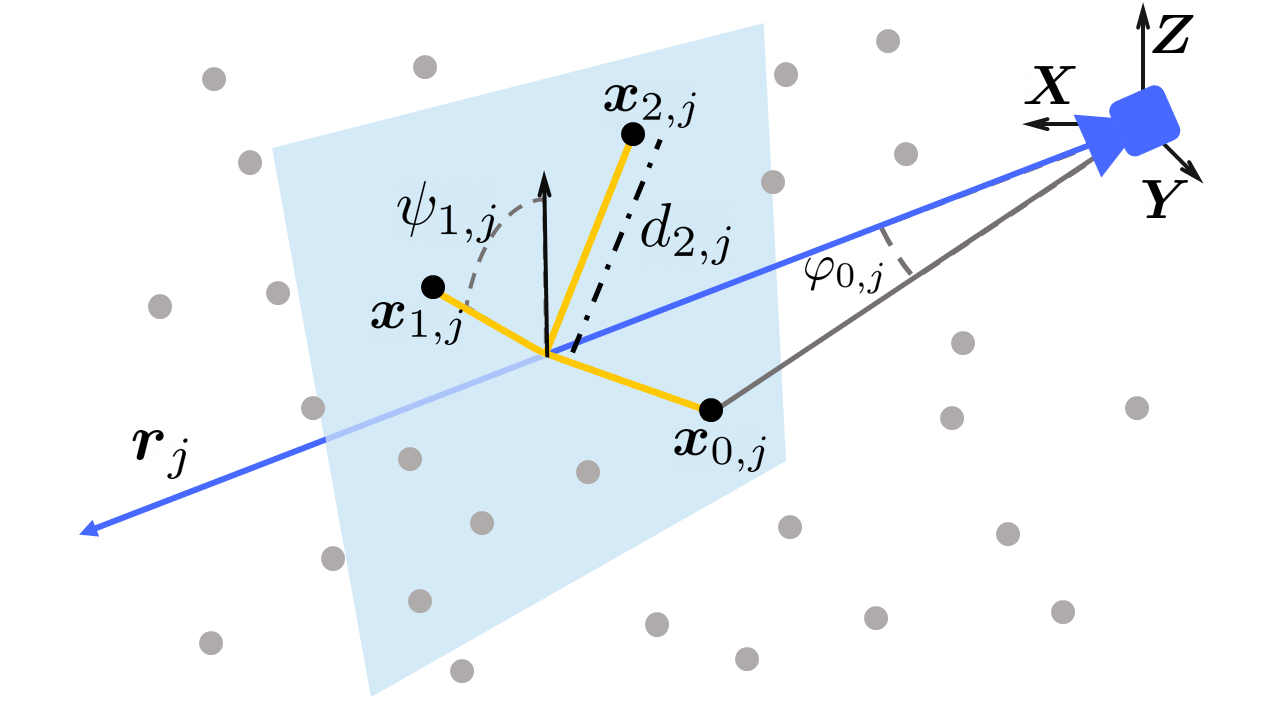}
	\caption{Ray-point distances are illustrated for a ray $j$ and the $k=3$ closest points. For better visualization, the ray and points are translated with $-\boldsymbol{o}_{j}$ into the scenes coordinate frame, and all points are projected into a single plane instead of 3 parallel planes.}\label{fig:pt_ray_d}
	\vspace{-12pt}
\end{figure}

As illustrated in Fig.~\ref{fig:pt_ray_d} and formalized in Eq.~\ref{eq:pt_ray_d}, \ref{eq:psi} and \ref{eq:theta}, we parameterize a close point 
using
% by
the angle $\theta_{k,j}$ between $\boldsymbol{x}_{k}$ and 
% the
ray $\boldsymbol{d}_{j}$, the orthogonal distance between the point and ray, and the angle $\psi$, 
% that is
defined as the radial coordinate of a projected $\boldsymbol{x}_k$ onto a plane 
determined by
% defined through
a projection of the global $\boldsymbol{Y}$-axis and it's cross product with the ray direction $\boldsymbol{d}_j$
% in Eq.~\ref{eq:project_x}.
%
\begin{equation}
\label{eq:project_x}
\begin{aligned}
    \begin{bmatrix}x \\ y \end{bmatrix}_{k,j,proj} =\begin{bmatrix} \boldsymbol{y}_{j}^T   \\ \left( \boldsymbol{d}_{j}  \times \boldsymbol{y}_{j} \right)^T \end{bmatrix} \boldsymbol{x}_{k}, &\\
        \text{with } \boldsymbol{y}_{j} = \frac{\boldsymbol{Y} - \left(\boldsymbol{Y} \cdot \boldsymbol{d}_{j}\right)\boldsymbol{d}_{j}}{\left\Vert \boldsymbol{Y} - \left(\boldsymbol{Y} \cdot \boldsymbol{d}_{j}\right)\boldsymbol{d}_{j} \right\Vert} \text{ and } \boldsymbol{y} \in \mathbb{R}^3, &
\end{aligned}
\end{equation}
\begin{equation}
        \psi_{k,j} = \arctan{\frac{x_{k,j,proj}}{y_{k,j,proj}}} .
    \label{eq:psi}
\end{equation}
The angle between the global point $\boldsymbol{x}_{k}$ and $\boldsymbol{d}_j$ is computed as
\vspace{-5pt}
\begin{equation}
    \theta_{k,j} = \arccos{\left(\boldsymbol{d}_{j,i} \cdot \frac{\boldsymbol{x}_{k}}{||\boldsymbol{x}_{k}||} \right)}.
    \label{eq:theta}
\end{equation}
This is computed in world coordinates, independently of local position, unlike $\varphi_{k,j}$ in Eq.~\ref{eq:phi}, that is used for computing the distance.
% Note that $\theta_{j,k}$ is computed between the global point $\boldsymbol{x}_{k}$ and $\boldsymbol{d}_j$ in world coordinates, independent of the position, and therefore different from the local $\varphi_{j,k}$, that computes the distance, that is
% %
% \vspace{-5pt}
% \begin{equation}
%     \theta_{k,j} = \arccos{\left(\boldsymbol{d}_{j,i} \cdot \frac{\boldsymbol{x}_{k}}{||\boldsymbol{x}_{k}||} \right)}.
%     \label{eq:theta}
% \end{equation}

\begin{figure}[bt!]
\vspace{-8pt}
	\centering
	\includegraphics[width=0.43\textwidth]{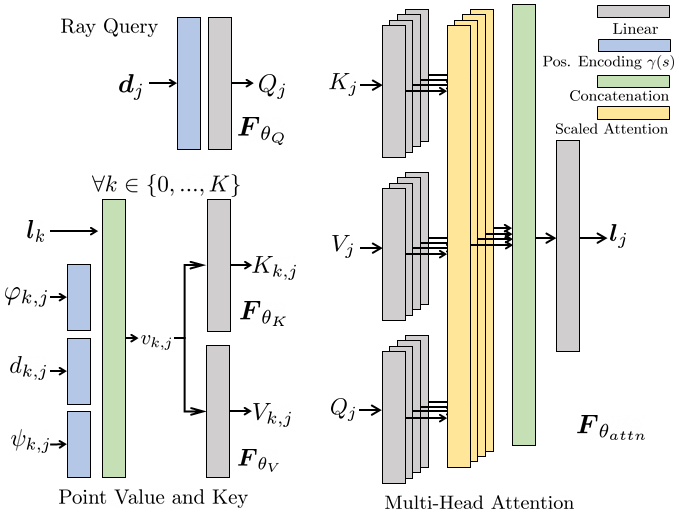}
	\vspace{-4pt}
	\caption{The multi-headed self-attention module aggregates the feature vector $\boldsymbol{l}_j$ of ray $j$ given the ray direction $\boldsymbol{d}_j$ from the information of the $K$ closest points.
	For each point $k$ an embedding $\boldsymbol{v}_{k, j}$ is computed from the point's feature and the 
	positionally encoded location relative
    to $\boldsymbol{r}_j$ for each ray-point pair $\left(j, k\right)$. $\boldsymbol{F_{\theta_{K}}}$ and $\boldsymbol{F_{\theta_{V}}}$ compute the  key $K$ and value $V$ vectors from $\boldsymbol{v}_{k, j}$. The query vector $Q$ is predicted for the ray direction $\boldsymbol{d}_j$.}
	\label{fig:enc_attn}
	\vspace{-8pt}
\end{figure}

%-------------------------------------------------------------------------
\vspace{0.5em}\noindent\textbf{Ray Feature Attention.}
Instead of applying an arbitrary
% chosen
weighting for the ray features, we propose a learned multi-head attention module 
(depicted in Fig.~\ref{fig:enc_attn})
% presented in Fig.~\ref{fig:enc_attn}
to compute 
% the
ray feature vector $\boldsymbol{l}_{j}$. We propose a variant of the multi-head attention module presented by Vaswani et al.~\cite{vaswani2017attention}. 
We compare the chosen attention based weighting with linear interpolation schemes In the experimental Sec.~\ref{sec:eval}.
% In the experimental Sec.~\ref{sec:eval} we compare the chosen attention based weighting with other linear interpolation schemes.
%
%
The two angular distances $\theta_{k, j}$ and $\psi_{k, j}$, as well as $d_{k, j}$ are transformed using a positional encoding \mbox{$\gamma\left(s\right) = \left[..., \sin{\left(2^t \pi s \right)}, \cos{\left(2^t \pi s \right)}, ...\right]$} with $t = 0,\ldots, T$
and $T=4$ ~\cite{mildenhall2020nerf, tancik2020fourier} to interpolate high frequency data from a low frequency input domain. The point 
feature vectors
% latents
$\boldsymbol{l}_k$ and the positional encoded distances are concatenated to form a unique descriptor \mbox{$\boldsymbol{v}_{k,j} = \left( \boldsymbol{l}_{k} \oplus \gamma\left(\theta_{k, j}  \right) \oplus \gamma\left(\psi_{k, j}  \right) \oplus \gamma\left(d_{k, j}  \right) \right)$}
corresponding to 
% between
ray $\boldsymbol{r}_j$ and point $k$, that encompasses the positional encoding and the feature vector of that point. The descriptor $\boldsymbol{v}_{k,j}$ is then passed through two
double-layer 
MLPs 
% with two layers
that predict a key $K_{k,j}$ and value $V_{k,j}$ for each of the $K$ point ray pairs. 
\vspace{-4pt}
\begin{equation}
    V_{k,j} = \boldsymbol{F}_{\theta_{V}}\left(\boldsymbol{v}_{k,j}\right)\text{, } K_{k,j} = \boldsymbol{F}_{\theta_{K}}\left(\boldsymbol{v}_{k,j}\right)
    \label{eq:value_key}
\end{equation}
\vspace{-9pt}
\begin{equation}
    Q_{j} = \boldsymbol{F}_{\theta_{Q}}\left(\gamma\left(\boldsymbol{d}_{j}\right)\right)
    \label{eq:query}
\end{equation}
The query vector $Q_j$ is derived 
from the positionally encoded ray direction $\gamma\left(\boldsymbol{d}_j\right)$.
% , we derive a query vector $Q_j$. The
The ray direction $\boldsymbol{d}_j$ is again represented in world coordinates
to make it independent of any local reference coordinate system.
% here, such that it will produce the same result independent from any coordinate reference in the locally captured scene.
The multi-head attention learns to predict a weight for all $V_{k,j}$ given $K_{k,j}$, for each selected point ray pair $\left(k, j\right)$ and 
% a
query ray $Q_{j}$. The aggregated output of the multi-head attention module 
comprises
% is then
a feature code $\boldsymbol{l}_j \in \mathcal{R}^{128}$, that describes the light fields for each ray $\boldsymbol{r}_j$:
% , that is
%
\vspace{-3pt}
\begin{equation}
    \text{multi-head attention: }\boldsymbol{l}_j =  \boldsymbol{F}_{\theta_{attn}}(  Q_{j}, K_{k,j}, V_{k,j}).
    \label{eq:mhattn}
\end{equation}

%-------------------------------------------------------------------------
\vspace{-6pt}
\vspace{0.0em}\noindent\textbf{Points Beyond the Point Cloud.}
Point clouds 
in most automotive datasets
% the captured point cloud data 
only capture the scene geometry from the ground plane up to a few meters height.
% of multiple meters. 
This results in scene regions
% Thus there are parts of the scene 
which are not explicitly captured in the point cloud data, such as high building structures and the sky. We therefore set a threshold $d_{\infty}$ below which we consider rays to intersect with the point cloud. The value $d_{\infty}$ is chosen as the maximum distance between two points in any $P_{i}$ after ignoring outlying points. 
For points
% Only for points
that exceed $d_{\infty}$, we concatenate $\boldsymbol{v}_{k,j}$ with a learned global feature code $\boldsymbol{l}_{\infty}$, such that the attention module can leverage both a global and
a local
point feature representation,
as point features may contain relevant context and geometry for structures that rise above the point cloud, and may therefore still be useful.
% . The point features are still useful as they may contain relevant context and geometry for structures that rise above the point cloud.

% \yb{I didn't comment the following:}
%\begin{equation}
%    \text{attn}(q_j, K_{k,j}, V_{k,j}) = \text{softmax} \left( \frac{q_j K_{k,j}^T}{\sqrt{d_k}}\right)V_{j,i}
%\end{equation}
%\begin{equation}
%    \boldsymbol{l}_j = attn( \{\boldsymbol{l}_0, ..., \boldsymbol{l}_K\}, \{\boldsymbol{x}_0, ..., \boldsymbol{x}_K\}, \boldsymbol{d}_j)
%\end{equation}

%
%-------------------------------------------------------------------------
\subsection{RGB Prediction}
After predicting
% Now that we have shown how we predict
a feature vector $\boldsymbol{l}_j$ for any ray $\boldsymbol{r}_j$ from encodings on a sparse point cloud, we are finally able to reconstruct the color $\boldsymbol{C}_j$ 
corresponding to any
% for an
arbitrary ray in our global scene, that is
\begin{equation}
    \boldsymbol{C}_j = \boldsymbol{F}_{\theta_{LF}} \left( \boldsymbol{d}_j \oplus \boldsymbol{l}_j\right).
    \label{eq:c_map}
\end{equation}
%
% Here we feed the $8$-layers MLP $\boldsymbol{F}_{\theta_{LF}}$ (with 256 channels) with the concatenation of ray direction $\boldsymbol{d}_j$ and  feature vector $\boldsymbol{l}_j$ corresponding to the ray at index $j$, to produce the predicted output color $\boldsymbol{C}_j$.
%Here, concatenating the ray direction $\boldsymbol{d}_j$ and the feature vector $\boldsymbol{l}_j$ corresponding to the ray at index $j$, we evaluate a learned color mapping function in Eq.~\ref{eq:c_map}. %, that predicts the color. %along each ray in the light field. 
% Ray direction and latent vector are concatenated $\left( \boldsymbol{d}_j \oplus \boldsymbol{l}_j\right)$ and the mapping from the light field representation is approximated with an eight layer and 256 wide MLP, that is trained jointly with all other modules.

Here $\boldsymbol{F}_{\theta_{LF}}$ is an $8$-layer MLP  (with 256 channels) that takes the concatenation of ray direction $\boldsymbol{d}_j$ and  feature vector $\boldsymbol{l}_j$ corresponding to the ray at index $j$, to predict output color $\boldsymbol{C}_j$. Implementation details for this and all other modules are provided in the supplementary materials.

For each predicted ray color $\hat{C}\left(\boldsymbol{r_j}\right)$ we can compute the mean-squared error image loss
\vspace{-8pt}
\begin{equation}
    \mathcal{L} = \sum_{j \in \mathcal{R}} \left\Vert \hat{C}\left(\boldsymbol{r_j}\right) - C\left(\boldsymbol{r_j}\right) \right\Vert_{2}^{2}.
    \label{eq:loss}
\end{equation}
%-------------------------------------------------------------------------
\vspace{0.5em}\noindent\textbf{Training}
All model parameters,
namely
% of the model, that is
$\theta_{ResNet18}$, $\theta_{K}$, $\theta_{V}$, $\theta_{Q}$, $\theta_{attn}$ and $\theta_{LF}$, are jointly optimized
by minimizing the loss in Eq.~\ref{eq:loss} using the Adam optimizer~\cite{Kingma2015AdamAM} with a linear learning rate decay, where
% using the respective gradients computed by back propagating from the presented loss in Eq.~\ref{eq:loss} to each parameter. 
at
% In
each step we randomly sample 8192 rays 
% for $\mathcal{R}$ 
from a small batch of frames.
% All parameter are trained using the Adam optimizer~\cite{Kingma2015AdamAM} with a linear learning rate decay.
	%------------------------------------------------------------------------
	% Method Sec. ...
	%------------------------------------------------------------------------
	\begin{figure*}[t!]
	\vspace{-10pt}
	\centering
	\resizebox{0.93\linewidth}{!}{
	\renewcommand{\arraystretch}{0.5}
	\begin{tabular}{@{}c@{\hskip .15cm}c@{\hskip .15cm}c@{\hskip .15cm}c@{\hskip .15cm}c@{}}
		\centering
		{\small RGB Frame}&
		{\small NeRF}&
		{\small DS-NeRF}&
		{\small GSN}&
		{\small Neural Point Light Fields}\\
		
		\includegraphics[width=.42\columnwidth, trim={0cm 0cm 0cm 0cm},clip]{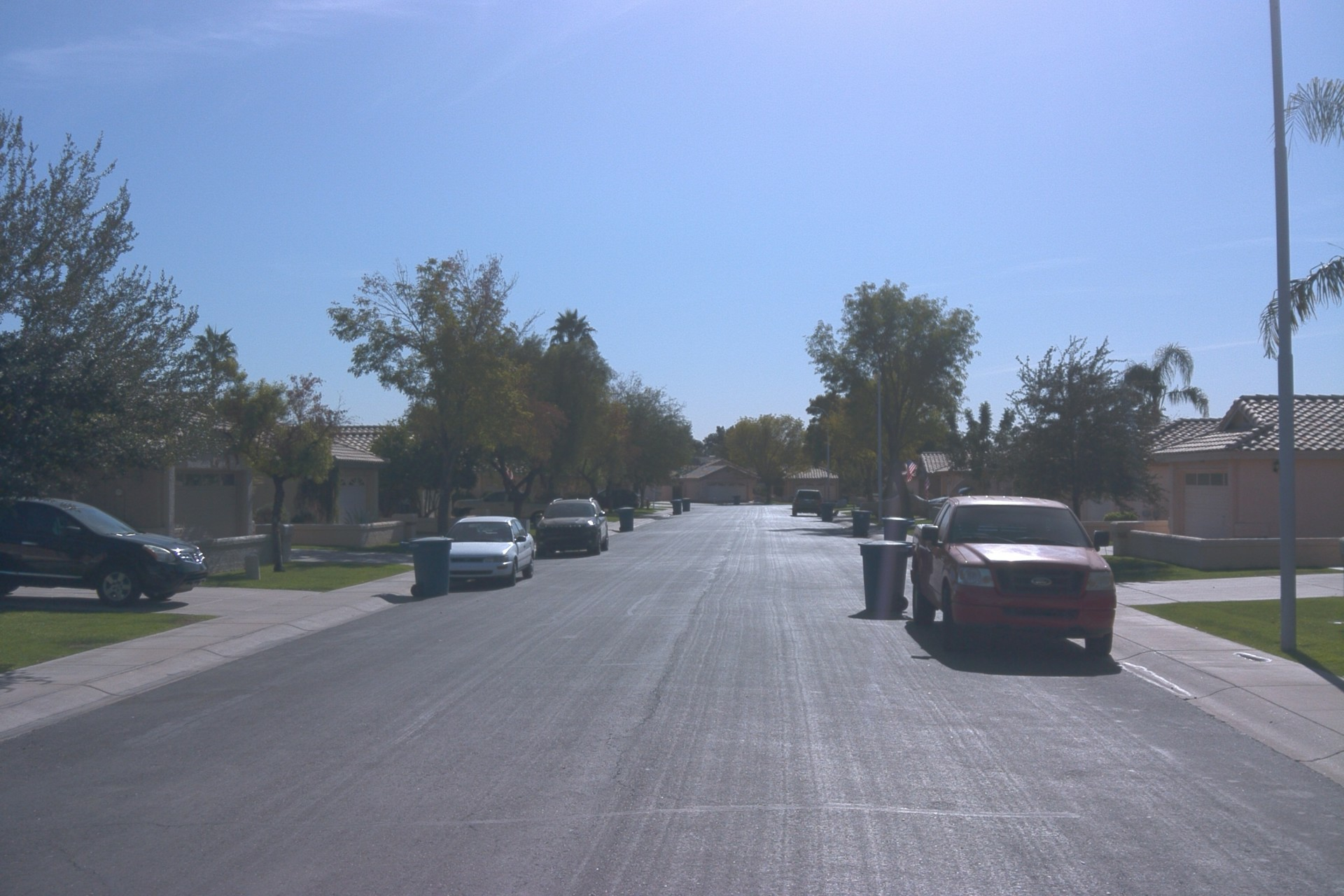}&
		\includegraphics[width=.42\columnwidth, trim={0cm 0cm 0cm 0cm},clip]{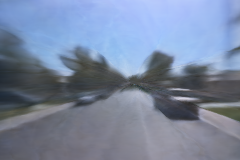}&
		\includegraphics[width=.42\columnwidth, trim={0cm 0cm 0cm 0cm},clip]{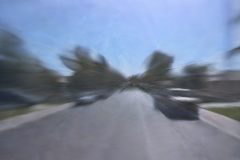}&
		\includegraphics[width=.28\columnwidth, trim={0cm 0cm 0cm 0cm},clip]{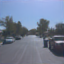}&
		\includegraphics[width=.42\columnwidth, trim={0cm 0cm 0cm 0cm},clip]{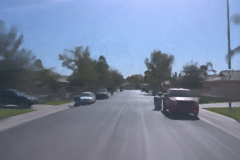}\\[.2cm]
		
		\includegraphics[width=.42\columnwidth, trim={0cm 0cm 0cm 0cm},clip]{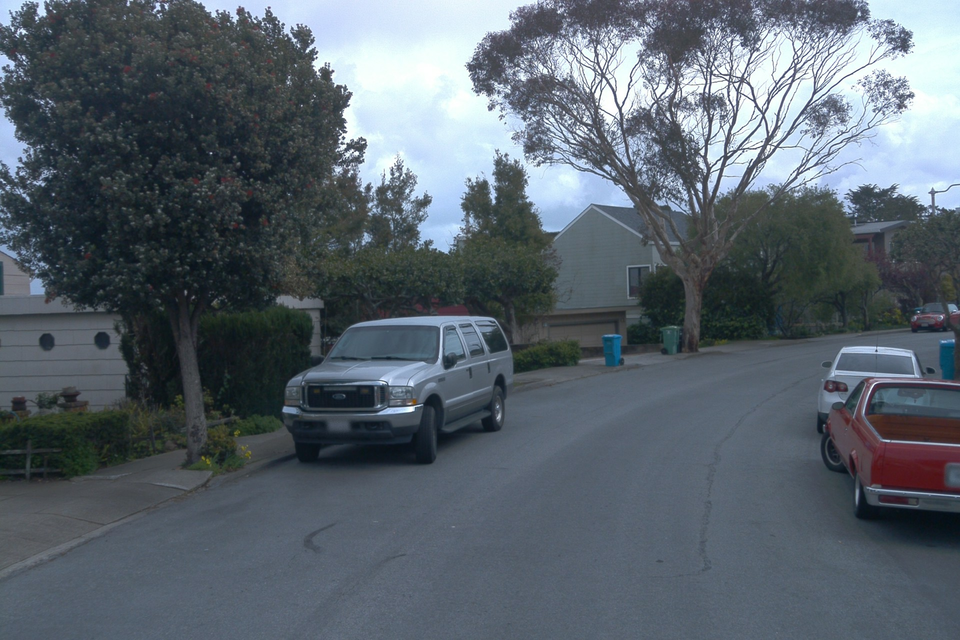}&
		\includegraphics[width=.42\columnwidth, trim={0cm 0cm 0cm 0cm},clip]{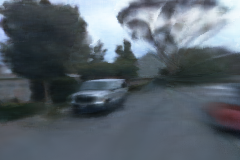}&
		\includegraphics[width=.42\columnwidth, trim={0cm 0cm 0cm 0cm},clip]{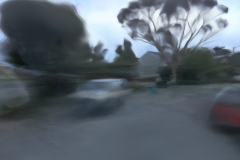}&
		\includegraphics[width=.28\columnwidth, trim={0cm 0cm 0cm 0cm},clip]{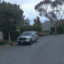}&
		\includegraphics[width=.42\columnwidth, trim={0cm 0cm 0cm 0cm},clip]{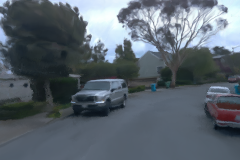}\\[.2cm]
		
		\includegraphics[width=.42\columnwidth, trim={0cm 0cm 0cm 0cm},clip]{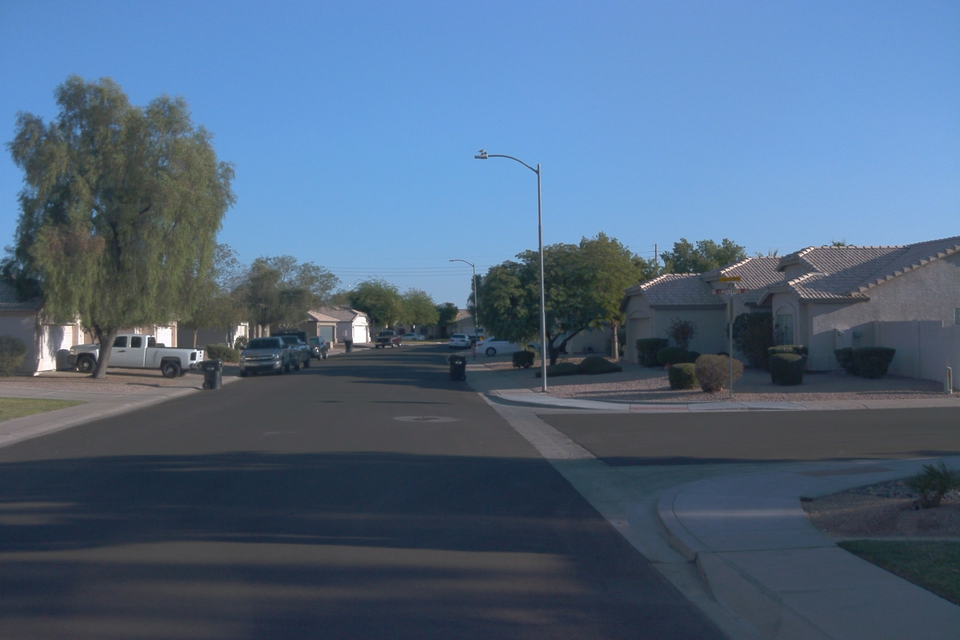}&
		\includegraphics[width=.42\columnwidth, trim={0cm 0cm 0cm 0cm},clip]{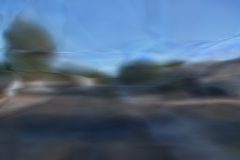}&
		\includegraphics[width=.42\columnwidth, trim={0cm 0cm 0cm 0cm},clip]{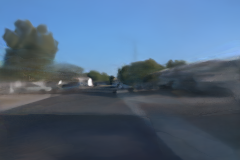}&
		\includegraphics[width=.28\columnwidth, trim={0cm 0cm 0cm 0cm},clip]{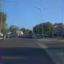}&
		\includegraphics[width=.42\columnwidth, trim={0cm 0cm 0cm 0cm},clip]{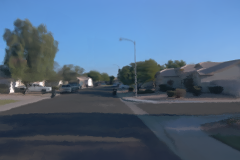}\\[.2cm]
		
	\end{tabular}
	}
	\vspace{-6pt}
	\caption{Scene Reconstruction. We present results for reconstructing images for poses seen during training of NeRF~\cite{mildenhall2019llff}, DS-NeRF~\cite{deng2021depth}, GSN~\cite{devries2021unconstrained} and Neural Point Light Fields. All methods were trained on the same set of scenes from the Waymo Open Dataset~\cite{sun2020scalability}. NeRF (even with substantially increased model capacity) and DS-NeRF show similar blurriness and other artifacts, while the depth supervision allows DS-NeRF to improve over existing methods. GSN produces fewer artifacts while struggling to reconstruct fine details, and fails for sparsely observed views (center scene). Neural Point Light Fields most faithfully reconstructs the image from the data set, see also Tab.~\ref{tab:quant_results}.}
	\label{fig:result_reconstruction}
 	\vspace{-12pt}
\end{figure*}

\section{Assessment}\label{sec:eval}
To assess the proposed method
and 
% , we 
evaluate its complexity,
we 
% and
train neural point light fields on an automotive driving dataset. 
% Specifically, 
We compare against state-of-the-art neural rendering methods by generating novel views interpolating between poses on the driven trajectory, as well as extrapolating to completely new trajectories. Moreover, we analyze how architecture and parameter choices in the proposed method affect reconstruction quality.
\subsection{Complexity}
Volumetric neural rendering methods require a large number of samples per ray for 
obtaining
accurate results. Even though existing methods allow speeding up rendering times~\cite{hedman2021baking}, training often requires hundreds of ray samples. We report the measured time and 
evaluations count
% counted evaluations
corresponding to processing a
single ray
% ray evaluation
during training and inference in Tab.~\ref{tab:complexity}.
To 
ignore differences related to specific implementation speed-ups (such as rays pre-caching),
% avoid differences due to pre-caching of rays or similar speed-ups in an implementation, the 
evaluation time is measured after the ray sampling step for a respective PyTorch~\cite{pytorch} implementation of the method. 
% The 
Measured times include encoding and decoding steps (e.g., point encoding in our method or convolution refinement in GSN), normalized by the number of image pixels to correspond to a single ray.
% normalized but divided by the number of pixels in an image to compute the time per ray. 

In contrast to volumetric scene representations, that need a high number of sampling points, even when supported by local feature vectors, Neural Point Light Fields only require a single evaluation per ray during rendering. 
This leads to a two times speedup, despite the overhead incured due to extraction of point features.

% \yb{I'm not sure I understand the next sentence, let's discuss}When measured, this results in two times speedup, despite our method incurs overhead due to extract point features.
\begin{table}[ht]
	\vspace{-1pt}
	\centering
	\resizebox{1.\columnwidth}{!}{%
		\renewcommand{\arraystretch}{0.9}
		\begin{tabular}[htb]{ll|cccc}
		Cost&\space
			& NeRF \cite{mildenhall2020nerf} 
			& DS-NeRF \cite{deng2021depth} 
			& GSN \cite{devries2021unconstrained}
			& Ours \\
			\hline
			\hline
			No. of Evaluations &$\downarrow$ & 192 & 192 & 64 & \textbf{1} \\
			Time per ray, training (in $\mu$s) &$\downarrow$ & 146  &  146 & \underline{37} & \textbf{34} \\
			Time per ray, inference (in $\mu$s) &$\downarrow$ & 49  &  49 & \underline{17} & \textbf{10} \\
		\end{tabular}
	}%
	% \vspace{-8pt}
	\caption{Complexity per ray during training and inference. All volumetric approaches require multiple evaluations per ray. Neural Point Light Fields (Ours) has a complexity of $O(1)$ per rendered ray. Despite an added complexity in the feature extraction step, this allows for shorter training and inference.}
	\label{tab:complexity}
	\vspace{-12pt}
\end{table}

\begin{figure*}[t!]
	\vspace{-10pt}
	\centering
	\resizebox{0.93\linewidth}{!}{
	\renewcommand{\arraystretch}{0.5}
	\begin{tabular}{@{}c@{\hskip .15cm}c@{\hskip .15cm}c@{\hskip .15cm}c@{\hskip .15cm}c@{}}
		\centering
		{\small RGB Frame}&
		{\small NeRF}&
		{\small DS-NeRF}&
		{\small GSN}&
		{\small Neural Point Light Fields}\\

		\includegraphics[width=.42\columnwidth, trim={0cm 0cm 0cm 0cm},clip]{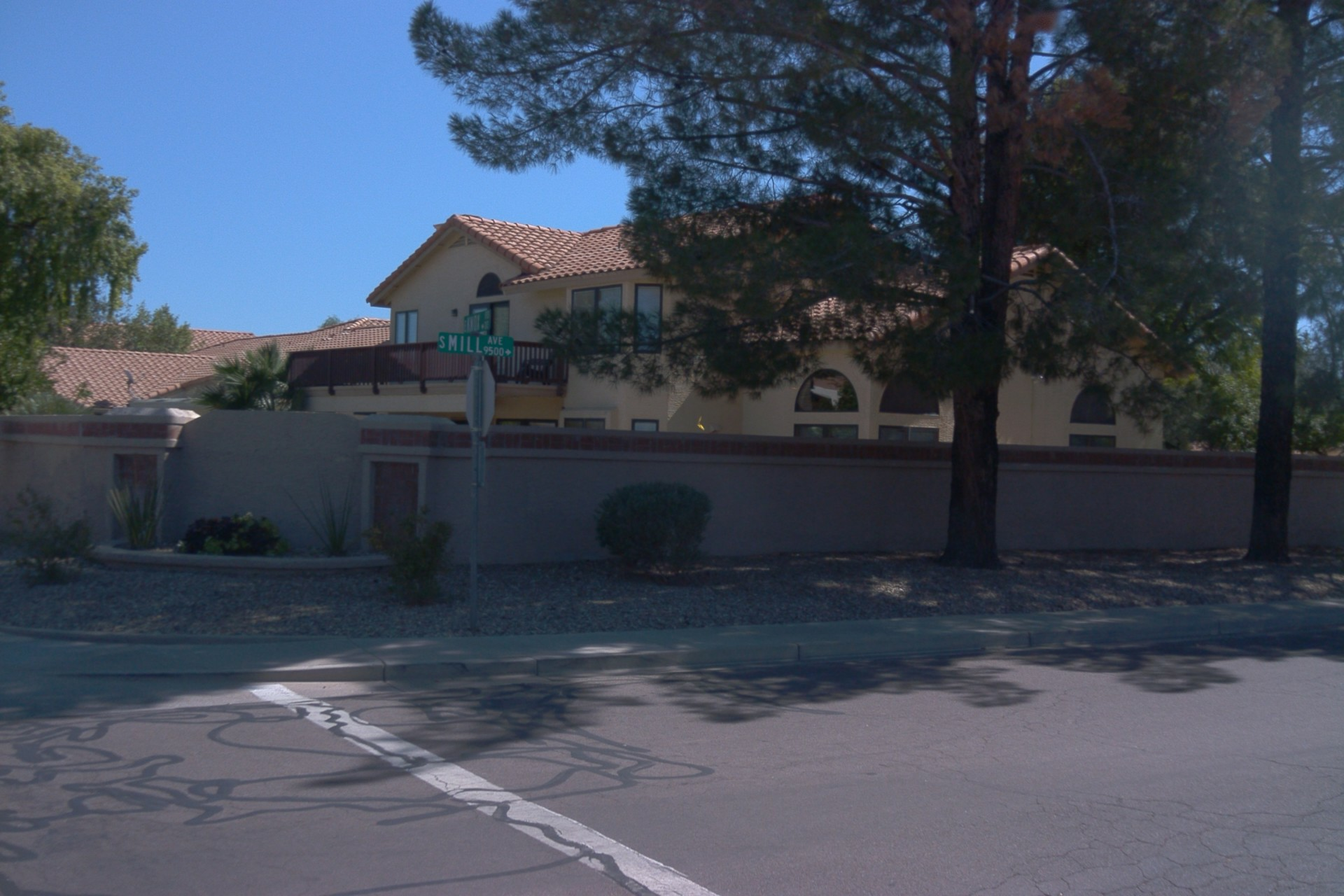}&
		\includegraphics[width=.42\columnwidth, trim={0cm 0cm 0cm 0cm},clip]{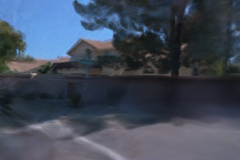}&
		\includegraphics[width=.42\columnwidth, trim={0cm 0cm 0cm 0cm},clip]{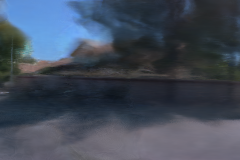}&
		% \parbox{0.37\columnwidth}{\textbf{N/A}}
		\includegraphics[width=.28\columnwidth, trim={0cm 0cm 0cm 0cm},clip]{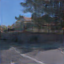}&
		\includegraphics[width=.42\columnwidth, trim={0cm 0cm 0cm 0cm},clip]{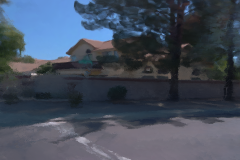}\\[.2cm]
		
		\includegraphics[width=.42\columnwidth, trim={0cm 0cm 0cm 0cm},clip]{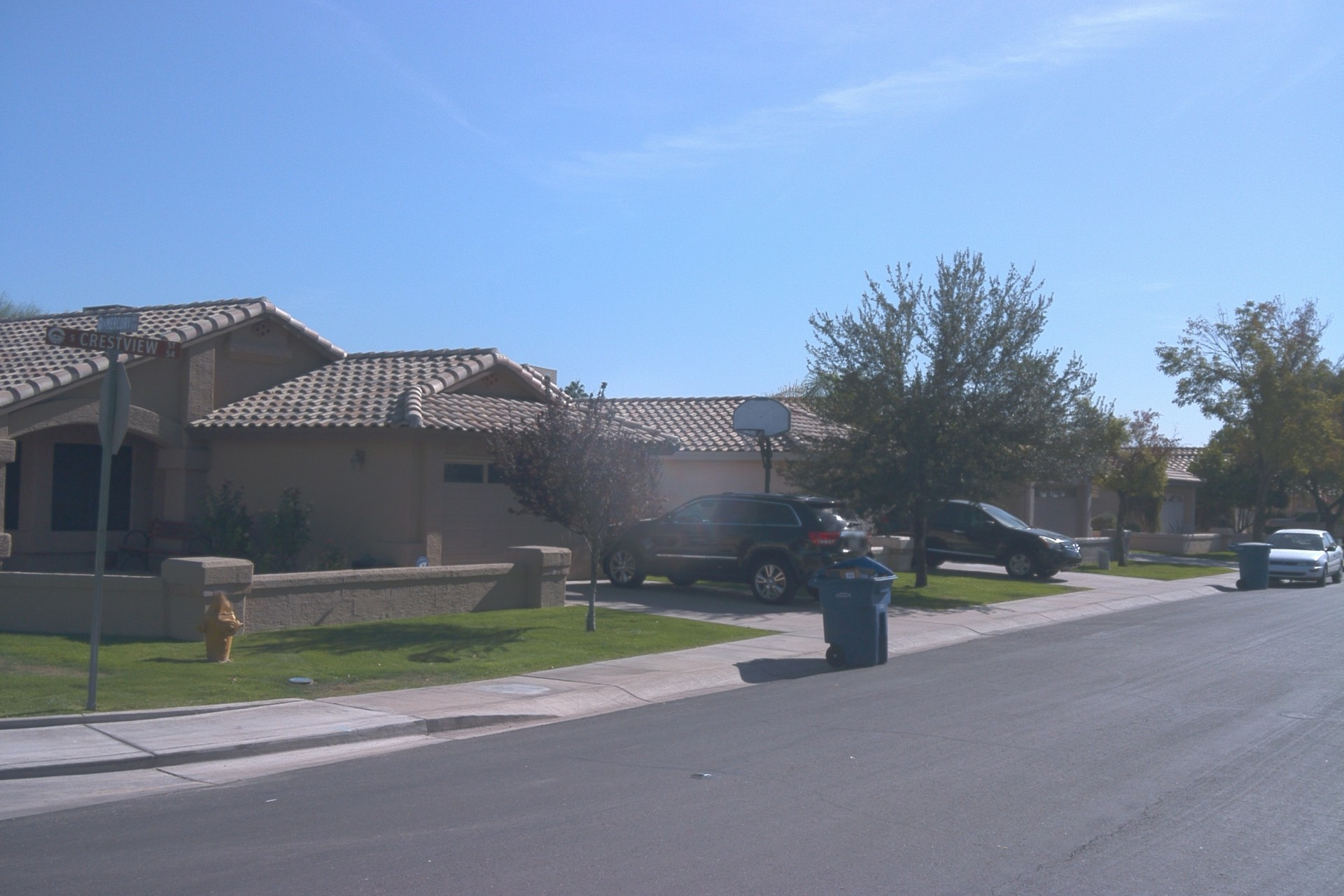}&
		\includegraphics[width=.42\columnwidth, trim={0cm 0cm 0cm 0cm},clip]{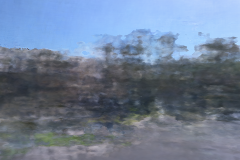}&
		\includegraphics[width=.42\columnwidth, trim={0cm 0cm 0cm 0cm},clip]{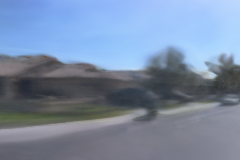}&
		\includegraphics[width=.28\columnwidth, trim={0cm 0cm 0cm 0cm}, clip]{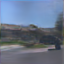}&
		\includegraphics[width=.42\columnwidth, trim={0cm 0cm 0cm 0cm},clip]{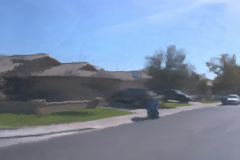}\\[.2cm]
				
		\includegraphics[width=.42\columnwidth, trim={0cm 0cm 0cm 0cm},clip]{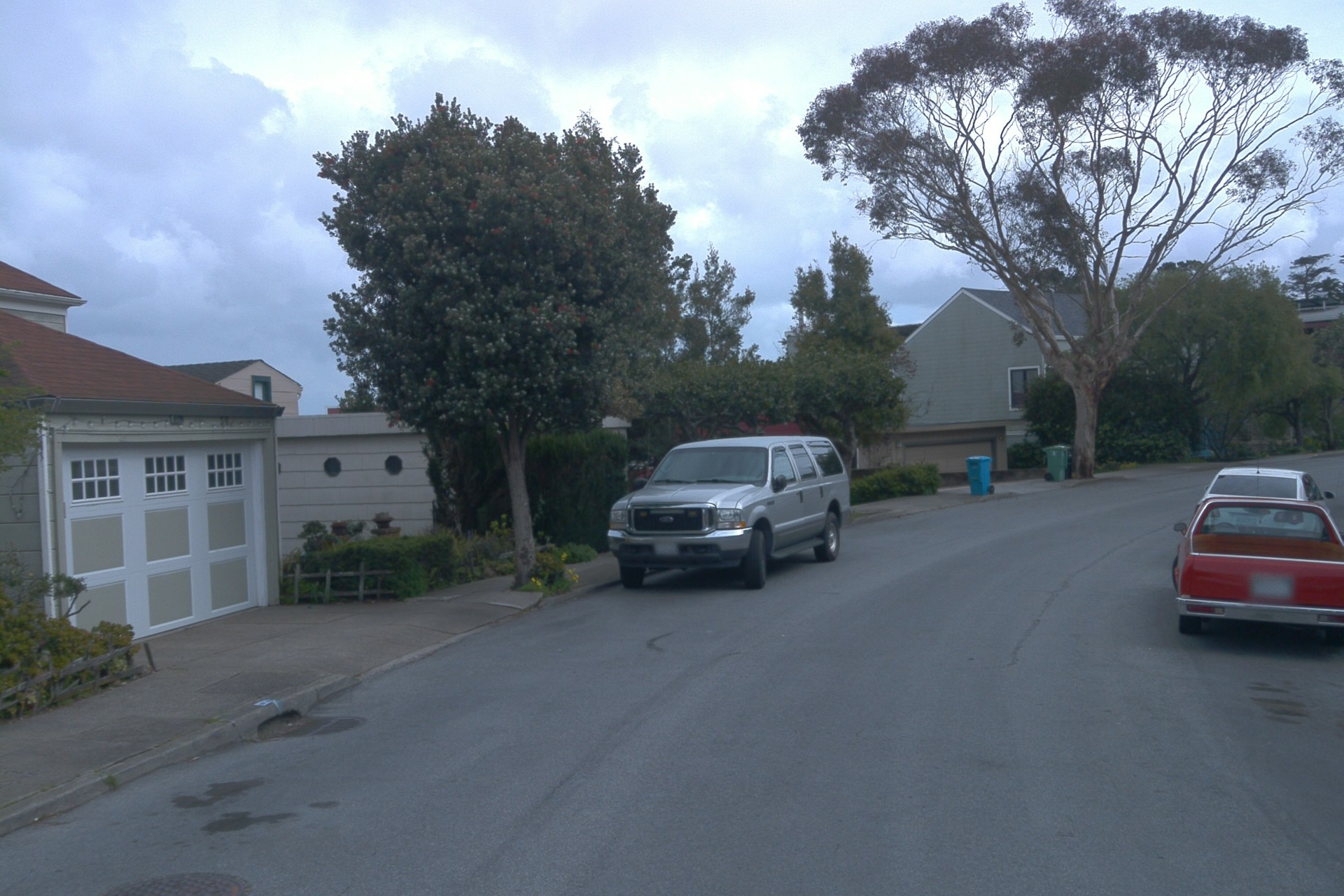}&
		\includegraphics[width=.42\columnwidth, trim={0cm 0cm 0cm 0cm},clip]{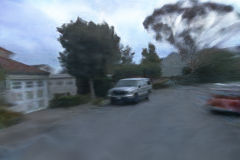}&
		\includegraphics[width=.42\columnwidth, trim={0cm 0cm 0cm 0cm},clip]{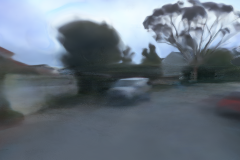}&
		\includegraphics[width=.28\columnwidth, trim={0cm 0cm 0cm 0cm},clip]{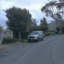}&
		\includegraphics[width=.42\columnwidth, trim={0cm 0cm 0cm 0cm},clip]{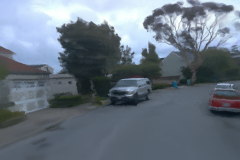}\\[.2cm]
		
	\end{tabular}
	}
% 	\vspace{-6pt}
	\caption{Novel View Interpolation. We predict views for unseen poses held-out from the training data.
    % 	for a scene. 
    images in middle row are taken
    % The images shown in the middle rows stem
    from the longest selected scenes (~200 frames), while the rest are taken from shorter ones (~80 frames). NeRF and DS-NeRF show blurry and overly smooth results, but perform better on smaller scenes. NeRF synthesizes the details on the small scenes better, while failing completely on larger scenes, even when substantially increasing the model's capacity. GSN performs consistently across all scenes, but exhibits artifacts and lacks detail. Our Neural Point Light Fields representation allows high-quality synthesis for novel view interpolation.}
	\label{fig:result_interpolation}
	\vspace{-8pt}
\end{figure*}
%
%-------------------------------------------------------------------------
\subsection{Experimental Setup}
We quantitatively and qualitatively validate the proposed method on two tasks, 
namely view 
reconstruction and novel view synthesis, where we compare against Generative Scene Networks (GSN), NeRF and depth-supervised NeRF  (DS-NeRF). GSN has been successfully applied to large scale indoor scenes \cite{devries2021unconstrained} and takes advantage of a local embedding of the scene that is jointly learned with the scene. In contrast to our sparse point features, the latent codes are located on a sparse 2D floorplan. We evaluate NeRF~\cite{mildenhall2020nerf} as a state-of-the-art volumetric scene representation. Additionally we evaluate DS-NeRF~\cite{deng2021depth}, which takes advantage of an additional depth supervision for the opacity prediction. \new{In the Supplementary Document, we present additional comparisons to NeRF++~\cite{zhang2020nerf++} and Free View Synthesis~\cite{riegler2020free}, which employs features on a mesh proxy geometry as discussed in Sec.~\ref{sec:related_work}.}
All methods were trained with their official publicly available code, by choosing the configuration closest to our outdoor/free moving scene scenario.
% the code, that was published by the authors and the experimental configurations closest to the scenario of outdoor/free moving scenes. 
For our method we use a maximum of $N=20000$ randomly sampled points, $K=8$ closest points, $128$ dimensional point and ray embedding $\boldsymbol{l}_k$ and $\boldsymbol{l}_j$, and $8$ heads in the multi-head attention module.

All methods except GSN were trained on 6 scenes from the Waymo Open Dataset \cite{sun2020scalability} with a length $\leq$ 200 frames, see Supplemental Document. To 
allow training
% train
on a single GPU, we downsample 
the captured images by a factor of $8$,
% to the eighth of the original resolution
resulting in a resolution of $240\times160$ pixels. For GSN, a convolutional refinement step requires the models to be trained on the full image resolution, and the code provided hard-coded settings that required us (after consulting with the authors) to
use $64\times64$ image crops.
% crop to a square resolution of 64x64.
For a fair evaluation, we report GSN results
% Results on GSN are provided
for 3 scenes,
while calculating metrics on 
% and all metrics for GSN are calculated on
downsampled dataset images.
% , to make the evaluation fair.
Note that GSN has an advantage in all quantitative evaluations as a smaller FOV at lower resolution needs to be synthesized. All models were trained until convergence on each scene on a mixture of NVIDIA TITAN Xp and NVIDIA V100 GPUs. Complexity evaluations were computed on the same hardware. The lower resolution requirements on GSN for over-fitting on a single scene resulted in a training time of 2 days, while the other models trained for 2 to 3 days,
depending on the number of scene frames.
% varying with scene frame count.

\begin{table}[t!]
	\vspace{-1pt}
	\centering
	\resizebox{1.\columnwidth}{!}{%
		\renewcommand{\arraystretch}{0.9}
		\begin{tabular}[htb]{l|cccc}
		\hline \hline
			& NeRF \cite{mildenhall2020nerf} 
			& DS-NeRF \cite{deng2021depth} 
			& GSN \cite{devries2021unconstrained}
			& Ours \\
			\hline
			\multicolumn{5}{c}{\small \textbf{Reconstruction}} \\
			% Reconstruction && \\
			\hline % \hline
			PSNR $\uparrow$ & \underline{29.48} & 26.53 & 17.98 & \textbf{31.52} \\
			SSIM \,$\uparrow$  & \underline{0.815}  &  0.778 & 0.512 & \textbf{0.882} \\
			LPIPS $\downarrow$ & 0.289 & 0.306 & \underline{0.136} & \textbf{0.110} \\
			
			\hline
			\multicolumn{5}{c}{\small \textbf{Novel View Synthesis}} \\
			% Novel Composition & & & \\
			\hline %\hline
			PSNR $\uparrow$ & 22.47 & \underline{26.15} & 16.83 & \textbf{29.96} \\
			SSIM \,$\uparrow$  & 0.700  &  \underline{0.772} & 0.464 & \textbf{0.868} \\
			LPIPS $\downarrow$ & 0.389 & 0.310 & \underline{0.174} & \textbf{0.119} \\
			\hline \hline
			
		\end{tabular}
	}%
	\vspace{-8pt}
	\caption{We report PSNR, SSIM and LPIPS on 5 static scenes from the Waymo Open Dataset \cite{sun2020scalability} using images from the front camera for NeRF \cite{mildenhall2020nerf}, 
	DS-NeRF \cite{deng2021depth}, GSN \cite{devries2021unconstrained} 
    % 	depth-supervised NeRF \cite{deng2021depth}, and generative    scene networks \cite{devries2021unconstrained} 
    and Neural Point Light Fields. For PSNR and SSIM, higher is better; for LPIPS lower is better. The best values are emphasized in \textbf{bold}, while the next best are \underline{underlined}. Our method outperforms all methods in all metrics. While NeRF shows only slightly worse reconstruction performance, DS-NeRF provides better novel view synthesis capabilities.}
	\label{tab:quant_results}
	\vspace{-12pt}
\end{table}
%-------------------------------------------------------------------------
\vspace{0.5em}\noindent\textbf{Quantitative Evaluation.}
%
%\yb{The next sentence already appears in the beginning of the section} We evaluate the proposed method for reconstruction of seen frames, and novel view synthesis of unseen, held-out frames, where we average evaluations\yb{metrics?} across all scenes.
We train all methods using the same $90\%$ of all driven trajectory frames,
leaving the remaining $10\%$ for evaluating interpolation of unseen views within the observed trajectory.
% The left out $10\%$ are used for evaluating the ability to interpolate novel views inside the seen trajectory. 
Tab.~\ref{tab:quant_results} reports quantitative results for both tasks using the PSNR, SSIM~\cite{wang2003ssim} and LPIPS~\cite{zhang2018perceptual} metrics. %\yb{Are you sure? NeRF gets 29.48, while we get 31.52. Anyway better rephrase the sentence:} While NeRF is close to the proposed method with respect to the PSNR metric on the reconstruction task it performs similar to DS-NERF on the SSIM and LPIPS metric. 
GSN makes an overall worse impression than the other methods in both tasks. The proposed method outperforms all other methods in all metrics. While NeRF performs significantly worse in the Novel View Synthesis task, DS-NeRF 
exhibit only a slight performance drop 
% shows only a slightly worse performance 
compared to its reconstruction results, probably benefiting from a better opacity prediction when trained on a sparse set of images. Our method performs the best in the view synthesis task as well, exhibiting only a minor performance degradation compared to the reconstruction task, in contrast to NeRF outputs.  

%-------------------------------------------------------------------------
\vspace{0.5em}\noindent\textbf{Scene Reconstruction.}
The results shown in Fig.~\ref{fig:result_reconstruction} support the quantitative evaluation from Tab.~\ref{tab:quant_results}.
While NeRF produces inconsistent and blurry predictions for the large scenes we address in this work, it is still able to recover some details on straight scenes. We hypothesize that the blurriness arises from the requirements of an accurate pose information and the sparse set of training views on 
long scene trajectories.
% a long trajectories through a scene. 
DS-NeRF shows a similar behavior, but lacks some detail that has been reconstructed in NeRF, while producing smooth artifacts. Renderings of the depth map of the trained scene suggest that the point cloud capture is too smooth for DS-NeRF representation and, as such, suppresses high frequency features. In contrast, GSN produces an overall consistent reconstruction, independent of scene length. Nevertheless results show smoothing even in the significantly downsampled resolution accepted by GSN. 
In contrast, Neural Point Fields allows reconstructing all structures, independent of their position and appearance across frames
resulting on only few artifacts on very fine structures (e.g., individual tree branches, leafs).
% . Few artifacts on very fine structures such as individual tree branches, leafs remain. 
Please also see the video in the Supplementary Material.

%-------------------------------------------------------------------------
\vspace{0.5em}\noindent\textbf{Novel View Trajectory Interpolation.}
We next compare views synthesized for frames excluded from the training data in Fig.~\ref{fig:result_interpolation}. DS-NeRF suffers from blur and ghosting in the interpolation task. NeRF shows similar, though slightly weaker artifacts on the few scenes it was able to converge on. 
Our method produces high-quality renderings when handling both short (top and bottom rows) and long (middle row) scenes.
% This stability issue for long scenes is \yb{lets rephrase:}evident, when we compare the results in the top and bottom scene, that are only half as long as the longer scene in the middle, that are only half as long as the two other scenes. 
The results validate that these existing methods are not able to effectively synthesize scenes just from a sparse set of images. GSN, which uses local support, 
seems to be more consistent, producing similar output quality in both tasks, regardless of scene length.
% is more consistent resulting in similar outputs in both tasks independent of scene length.
Neural Point Light Fields encode the scenes features on a sparse set of points, hence achieve high-quality novel view interpolation, even for long sequences.

\begin{figure}[t!]
	\vspace{-6pt}
	\renewcommand{\arraystretch}{0.5}
	\centering
	\begin{tabular}{@{}c@{\hskip .05cm}c@{\hskip .05cm}c@{\hskip .05cm}c@{\hskip .05cm}c@{\hskip .05cm}}
	    Trajectory&
	    NeRF&
	    DS-NeRF&
    	GSN&
    	Ours\\
	    \includegraphics[width=.21\columnwidth, trim={2cm .8cm 0cm 1cm},clip]{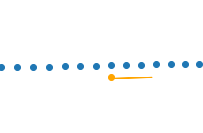}&
	    \includegraphics[width=.21\columnwidth, trim={0cm 0cm 0cm 0cm},clip]{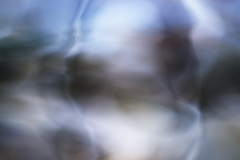}&
	    \includegraphics[width=.21\columnwidth, trim={0cm 0cm 0cm 0cm},clip]{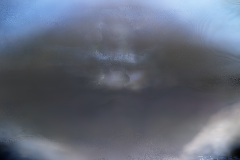}&
	    \includegraphics[width=.14\columnwidth, trim={0cm 0cm 0cm 0cm},clip]{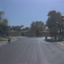}&
	    \includegraphics[width=.21\columnwidth, trim={0cm 0cm 0cm 0cm},clip]{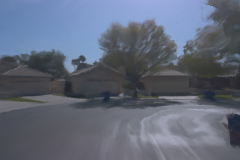}\\
	    
	    \includegraphics[width=.21\columnwidth, trim={0cm 0cm 0cm 0cm},clip]{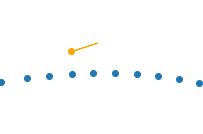}&
	    \includegraphics[width=.21\columnwidth, trim={0cm 0cm 0cm 0cm},clip]{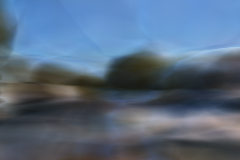}&
	    \includegraphics[width=.21\columnwidth, trim={0cm 0cm 0cm 0cm},clip]{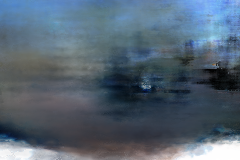}&
	    \includegraphics[width=.14\columnwidth, trim={0cm 0cm 0cm 0cm},clip]{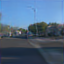}&
	    \includegraphics[width=.21\columnwidth, trim={0cm 0cm 0cm 0cm},clip]{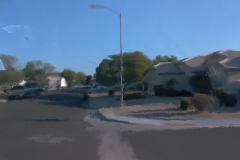}\\
    \end{tabular}
	\vspace{-6pt}
	\caption{Novel View Trajectory Extrapolation. Extrapolating views (orange) using the training trajectory (blue). While NeRF and DS-NeRF fail to synthesize views far from the training trajectory, the proposed method produces high quality results, similar to its performance in the reconstruction and view interpolation tasks.
% 	maintain a similar quality to  produces similar quality when compared to reconstruction and trajectory interpolation.
	}
	\label{fig:result_extrapolation}
	\vspace{-10pt}
\end{figure}

%-------------------------------------------------------------------------
\vspace{0.5em}\noindent\textbf{Novel View Trajectory Extrapolation.}
The results shown in Fig.~\ref{fig:result_extrapolation} report visual extrapolation experiments. We present a map of the novel 
view camera poses
% views camera pose
with respect to the training trajectory. Our method is able to generate a set of novel trajectories and scenes, that can hardly be differentiated from the interpolation and reconstruction results. 
This is possible 
within certain regions of the scene which are at least partially covered by the training images, see Supplemental Material. Views of scene regions which were not seen during training, e.g., the back of a vehicle only seen from the front, result in imaginary objects,
probably hallucinated from points similar to the observed objects.
% from conditioned on points similar to seen objects. 
Incorporating information from additional cameras (possibly covering $360^\circ$) may allow synthesizing such occluded scene regions in the future.
% In the future, including additional cameras in a 360 surround view application, may allow to cover even such entirely occluded scene parts.

\vspace{-0pt}
\subsection{Ablation Experiments}
\begin{figure}[t!]
	\vspace{-12pt}
	\centering
	\resizebox{0.92\columnwidth}{!}{%
	\begin{tabular}{@{}c@{\hskip .1cm}c@{\hskip .2cm}c@{}}
	    {\small Heuristic} & {\small $K=0$} & {\small $K=1$} \\
	    
    	\includegraphics[width=.379\columnwidth, trim={0cm 0cm 0cm 0cm},clip]{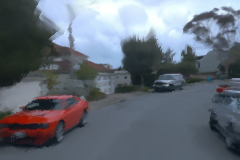}&
    	\includegraphics[width=.309\columnwidth, trim={4cm 0.5cm 0cm 1.5cm},clip]{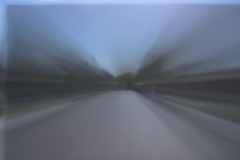}&
    	\includegraphics[width=.309\columnwidth, trim={4cm 0.5cm 0cm 1.5cm},clip]{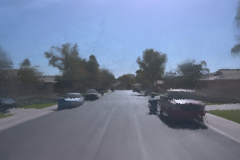}\\
    	
    	{\small Self-Attn. (Ours)} & {\small $K=2$} & {\small $K=8$ (Ours) } \\
    	
    	\includegraphics[width=.379\columnwidth, trim={0cm 0cm 0cm 0cm},clip]{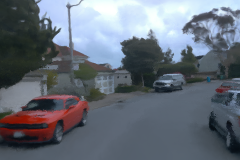}&
        \includegraphics[width=.309\columnwidth, trim={4cm 0.5cm 0cm 1.5cm},clip]{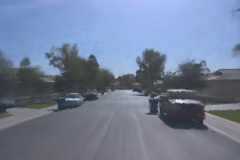}&
    	\includegraphics[width=.309\columnwidth, trim={4cm 0.5cm 0cm 1.5cm},clip]{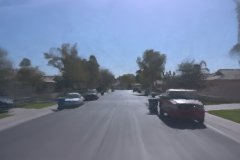}\\
    \end{tabular}
		}
		\vspace{-3pt}
    	\resizebox{1.\columnwidth}{!}{%
		\renewcommand{\arraystretch}{1.3}
		\begin{tabular}[htb]{l|cccccc}
		\hline \hline
			& Naive Sum 
			& Heuristic  
			& $K=0$ 
			& $K=1$ 
			& $K=2$ 
			& Ours \\
			% Reconstruction && \\
			\hline %\hline 
			PSNR $\uparrow$ & 4.84 & 24.56 & 18.88 & 29.83 & 30.95 & \textbf{31.52} \\
			\hline \hline
		\end{tabular}
	}%
	\vspace{-6pt}
	\caption{Ablation studies. Qualitative and quantitative
	comparisons of using different numbers $K$ of closest points per ray and different feature aggregation approaches.
    % 	results of ablations for the number of closest points $K$ and different approaches for the light field interpolation module.
	}
	\label{fig:ablations}
	\vspace{-8pt}
\end{figure}
We analyze architecture and parameter choices in Fig.~\ref{fig:ablations}. 
Choosing self-attention for aggregating ray features proves to be crucial, as we find
% The proposed self-attention to aggregate a single ray feature is crucial. We found 
that a heuristic weighting or naive summation over all point features are not able to achieve similar results. While 
merely summing prohibits training at all,
% the naive sum is not able to train at all, a 
heuristic weighting of each point feature by the inverse distance $d_{k,j}$ achieves better results. However, this weighting still lacks details and suffers from artifacts and noisy scene reconstruction.
We propose to index a set of points, in contrast to methods that purely parameterize a ray. 
In addition, we compare results for different number of points $K$ per ray in Fig.~\ref{fig:ablations}, indicating that a handful of points is essential for learning large scene light fields.
% Also the decision to use a set of points is necessary to learn the light field of such a large scene, that can not be embedded solely on a single or two points. We choose $K={0,1,2,8}$ to illustrate this effect in Fig.~\ref{fig:ablations}. Further
Additional ablation studies are reported in the Supplementary Materials.

	\section{Conclusion}\label{sec:conclusion}
We introduce an implicit representation that encodes a local light field on a point cloud. Departing from volumetric representations that require querying radiance estimates at hundreds of sample points along
each ray,
% a ray cast into the volume, 
we learn realistic radiance fields with only a single radiance sample per ray. Neural point light fields are functions of the ray direction and local point feature neighborhood, which allows us to interpolate the light field conditioned training images without densely captured input views. As such, the method allows for novel view synthesis in large-scale automotive scenarios, with only a few sparse view directions available during a drive-by capture. We validate the proposed method for novel view synthesis when interpolating and extrapolating along unseen trajectories, where existing implicit representation methods fail. While it is typical in automotive scenarios to have point cloud captures available, in the future, we plan to jointly recover point positions and local features of the proposed neural point light fields.

	%------------------------------------------------------------------------
	\vspace{6pt}
	
	\noindent \textbf{Acknowledgements.}
	We thank ServiceNow for providing compute resources for this project with the ServiceNow Toolkit. Felix Heide was supported by an NSF CAREER Award (2047359), a Sony Young Faculty Award, and a Project X Innovation Award. 
	Yuval Bahat was supported by the MSCA COFUND STAR fellowship.
	{\small
		\bibliographystyle{ieee_fullname}
		\bibliography{bib}
	}
\end{document}